\documentclass{article} 

\usepackage{graphicx} 
\usepackage{subfigure} 
\usepackage{amssymb}
\usepackage{amsmath}
\usepackage{wrapfig}
\usepackage{natbib}
\usepackage{algorithm}
\usepackage{algorithmic}
 \allowdisplaybreaks[1]
\def\V{v}
\newcommand{\SSS}{\mathcal{S}}
\def\GTDl{GTD($\lambda$)}

\title{Scaling Life-long Off-policy Learning}

\author{Adam White, Joseph Modayil, and Richard S.\ Sutton\\\\ Reinforcement Learning and Artificial Intelligence Laboratory\\
Department of Computing Science, University of Alberta}

\begin{document}
\maketitle

\begin{abstract}

We pursue a life-long learning approach to artificial intelligence that makes extensive use of reinforcement learning algorithms. We build on our prior work with general value functions (GVFs) and the Horde architecture. GVFs have been shown able to represent a wide variety of facts about the world's dynamics that may be useful to a long-lived agent (Sutton et al.\ 2011). We have also previously shown scaling---that thousands of on-policy GVFs can be learned accurately in real-time on a mobile robot (Modayil, White \& Sutton 2011). That work was limited in that it learned about only one policy at a time, whereas the greatest potential benefits of life-long learning come from learning about many policies in parallel, as we explore in this paper. Many new challenges arise in this off-policy learning setting. To deal with convergence and efficiency challenges, we utilize the recently introduced GTD($\lambda$) algorithm. We show that \GTDl ~with tile coding can simultaneously learn hundreds of predictions for five simple target policies while following a single random behavior policy, assessing accuracy with interspersed on-policy tests. To escape the need for the tests, which preclude further scaling, we introduce and empirically validate two online estimators of the off-policy objective (MSPBE). Finally, we use the more efficient of the two estimators to demonstrate off-policy learning at scale---the learning of value functions for one thousand policies in real time on a physical robot. This ability constitutes a significant step towards scaling life-long off-policy learning.
\end{abstract} 

%\twocolumn

%Arguably, the computational constraints of this big-data problem make other approaches, such as batch and least-squares approaches, inappropriate for life-long learning. 
%In this paper we focus on this key next challenge for life-long learning: learning efficiently about multiple target policies at once, while following a single behavior policy. 
%\section{Big data in robotics}
\section{Introduction}

Life-long learning is an approach to artificial intelligence based on learning from a long-stream of sensorimotor interaction generated by an agent interacting with its environment, Life-long learning emphasizes continual learning by an autonomous agent over long periods of time, perhaps months or years. This big data problem requires algorithms that scale efficiently to learn a multitude of diverse facts about the large stream of sensorimotor data. We purse a novel approach to knowledge acquisition and verification based on algorithms from reinforcement learning and a life-time of sensorimotor interaction of a mobile robot.  

General value functions (GVFs) provide an expressive language for representing sensorimotor knowledge about a long-lived agent's interaction with the world (Sutton et al.\ 2011). Knowledge is represented as approximate value functions with a reward function, an outcome function and a pseudo-termination function conditioned on a  policy. GVFs provide a semantics for experiential knowledge that is grounded in the sensorimotor data and verifiable by the agent, without human intervention---essential for scalability. 

In recent computational studies (Modayil, White \& Sutton 2011) have shown that predictions represented as GVFs can be learned at a massive scale with a high degree of accuracy. Under on-policy sampling, a mobile robot learned thousands of predictions about future sensor readings and state variables at several time scales operating at a 100 ms time step. These predictions were shown to be accurate compared to the optimal off-line solution. 

Previous work focused on a limited form of prediction, learning the consequences of a single policy. The greatest potential benefit of life-long learning comes from learning about many policies in parallel, using off-policy learning. Parallel learning introduces a new dimension of scaling not considered in typical sequential life-long learning systems and has yet to be demonstrated at scale and in real time on a robot.

Many new challenges arise in this off-policy learning setting. Scaling life-long learning in this way requires efficient learning methods that are robust under off-policy sampling. We use recently developed gradient temporal-difference learning method, \GTDl, with linear function approximation, to learn GVFs on a robot. Gradient TD methods are the only learning methods that scale linearly in the size of the feature set, require constant computation time per step, do not require memory or a forgetting process, and are therefore the only methods available for off-policy life-long learning at scale on a mobile robot. Arguably, the computational constraints of this big-data problem make batch and least-squares approaches inappropriate for life-long learning, while recent work in machine learning has highlighted the value of simple online learning methods for big data problems (see Hsu et al., 2011).       

Evaluating off-policy learning at scale poses an additional challenge: determining prediction accuracy for policies that are never executed by the robot. We first show that our robot can learn hundreds of predictions about several policies, with interspersed on-policy tests. However, these tests require interrupting learning, placing an upper bound on the number of policies the robot can learn about. We propose two efficiently computable, online measures of off-policy learning progress based on the off-policy objection function (MSPBE). Using these online measures, we demonstrate learning a thousand GVFs about a thousand unique target policies in real time on a robot. Our results represent a significant step towards scaling life-long learning.

\section{On-policy and off-policy prediction with value functions}
To begin, we consider how the problem of prediction is conventionally formulated in reinforcement learning.  The interaction between an agent and its environment is modelled as a discrete-time dynamical system with function approximation.  On each discrete time step $t$, the agent observes a feature vector $\phi_t \in \Phi=\mathbb{R}^n$, that partially characterizes the current state of the environment. Note that the agent has no access to the underlying environment state space $S$, or to the current state $s_t\in \SSS$; the observed feature vector, $\phi_t$, is computed from information available to the agent and thus is only implicitly a function of the environmental state, $\phi_t= \phi(s_t)$.  At each time step, the agent takes an action $a_{t} \in \mathcal{A}$,
%, according to a behaviour policy $b: \Phi \times \mathcal{A} \rightarrow [0,1]$, 
and the environment transitions into a new state producing a new feature vector, $\phi_{t+1}$.

In conventional reinforcement learning, we seek to predict at each time the total future discounted reward, where {\em reward} $r_t\in \mathbb{R}$ is a special signal received from the environment.  More formally, we seek to learn a value function $V: S \rightarrow \mathbb{R}$, conditional on the agent following a particular policy. The time scale of the prediction is controlled by a discount factor $\gamma \in [0,1)$. With these terms defined, the precise quantity being predicted, called the {\em return} $g_t \in \mathbb{R}$, is 
$$ g_t = \sum_{k=0}^\infty \gamma^k r_{t+k+1}, $$
and the value function is the expected value of the return,
$$V(s) = \mathbb{E}_\pi \left[\sum_{k=0}^\infty \gamma^k r_{t+k+1} \Big|  s_t=s \right],$$
where the expectation is conditional on the actions (after $t$) being selected according to a particular policy $\pi: \Phi \times \mathcal{A} \rightarrow [0,1]$.
As is common in reinforcement learning, we estimate $V$ with a linear approximation, $V_\theta(s) = \theta^\top \phi(s) \approx V(s)$, where $\theta \in \mathbb{R}^n$. 
%Since we are learning online, the value on time step $t$ can be approximated using the current weight vector $\theta_t \in \mathbb{R}^k$, $\V_t = \theta^T_t\phi_t.$, where $\theta_t \rightarrow \theta$ as $t \rightarrow \infty$. 

In the most common {\em on-policy} setting, the policy that conditions the value function, $\pi$, is also the policy used to select actions and generate the training data.
In general, however, these two policies may be different.
The policy that conditions the value function is called the {\em target} policy because it is the target of the learning process, and in this paper we will uniformly denote it as $\pi$.
The policy that generates actions and behaviour is called the {\em behaviour} policy, and in this paper we will denote it as $b: \Phi \times \mathcal{A} \rightarrow [0,1]$.
The most common setting, in which the two policies are the same, is called {\em on-policy learning}, and the setting in which they are different is called {\em off-policy learning}.
 
Conventional algorithms such as TD$(\lambda)$ and Q-learning can be applied with function approximation in an on-policy setting, but may become unstable in an off-policy setting. 
%Several algorithms can estimate the above expectation when the agent is following the target policy, $\pi$ (on-policy).  
Fewer algorithms work reliably in the off-policy setting.
%, in which the agent instead follows a behaviour policy, $b$, that is different from the target policy, $\pi$.
%Learning $\hat \V$ in this way is said to be off-policy because the expectation is conditioned on a policy different from the policy used to control the system.
One such algorithm is GTD($\lambda$), a gradient-TD algorithm designed to learn from off-policy sampling with function approximation (Maei, 2011). GTD($\lambda$) is an incremental prediction algorithm, similar to TD($\lambda$) (Sutton, 1988),  except with an additional secondary set of learned weights $w$, and an additional step size parameter $\alpha_w$. The algorithm retains the computational advantages of TD($\lambda$): its computational complexity is $\mathcal{O}(n)$ per step, and it can operate online and in real time. Unlike TD($\lambda$), GTD($\lambda$) is guaranteed to converge under off-policy sampling and with function approximation (linear and non-linear).  The following pseudocode specifies the GTD($\lambda$) algorithm.\\ 
\hrule
\begin{algorithmic}
\STATE Initialize $w_0$ and $e_0$ to zero and $\theta_0$ arbitrarily.
%\PRINT Initialize $e=\vec 0$ \\
%\PRINT Let $A_t$ be from $S_t$ according to $\pi_b$, and arrive at $S_{t+1}$. \\
%Observe sample, ($x_t$, $R_{t+1}$, $Z_{t+1}$, $x_{t+1}$), where $x_t = x(S_t)$. \\
\FOR {each time step $t$, given observed  sample $\phi_t, a_t, \phi_{t+1}$, and $r_{t+1}$} 
\STATE $\delta_t \leftarrow r_{t+1} + \gamma\theta_t^\top \phi_{t+1}   - \theta_t^\top \phi_t$
\STATE $\rho_t \leftarrow \frac{\pi(a_t|\phi_t) }{ b(a_t|\phi_t)}$
\STATE $e_t \leftarrow \rho_t (\phi_t + \gamma \lambda e_{t-1})$ 
\STATE $\theta_{t+1} \leftarrow \theta_t + \alpha_v ( \delta_t e_t - \gamma(1 - \lambda)(e^\top_t w_t)\phi_{t+1}) $ 
\STATE $w_{t+1} \leftarrow w_t + \alpha_w ( \delta_t e_t - (\phi_t^\top w_t) \phi_t)$ 
\ENDFOR
\label{GTD}
 \end{algorithmic}
\hrule
\vspace*{4mm}

The GTD($\lambda$), algorithm minimizes the $\lambda$-weighted mean-square projected Bellman error
\begin{equation} \textrm{MSPBE}(\theta,\Phi)=||V_\theta - \Pi_\Phi T_\pi^{\lambda,\gamma}V_\theta ||^2_D \end{equation} 
where $\Phi$ is the  matrix of all possible feature vectors $\phi$, $\Pi_\Phi$ is a projection matrix that projects the value function onto the space representable by $\Phi$, $T_\pi^{\lambda,\gamma}$ is the $\lambda$-weighted Bellman operator for the target policy $\pi$ and discount factor $\gamma$, and $D$ is a diagonal matrix whose diagonal entries correspond to the state visitation frequency under the behaviour policy $b$. %$$
%\begin{array}{rl}
%       \delta_t =& r_{t+1} + (1 - \gamma_{t+1})z_{t+1} + \gamma_{t+1}\theta_t^\top x_{t+1}   - \theta_t^\top x_t \\
%       \rho_t =& \frac{\pi(A_t|S_t) }{ \pi_b(A_t|S_t)}\\
%       e_t =& \rho_t (x_t + \gamma_t \lambda_t e_{t-1}) \\
%       \theta_{t+1} =& \theta_t + \alpha ( \delta_t e_t - \gamma_{t+1}(1 - \lambda_{t+1})(e^\top_t w_t)x_{t+1})  \\
%       w_{t+1} =& w_t + \beta ( \delta_t e_t - (x_t^\top w_t) x_t)
%     \end{array}
%$$

\section{An architecture for large-scale, real-time off-policy learning on robots}

%In this section, we describe how off-policy learning methods scale to learn from big data in real time on a robot, using an architecture introduced by Sutton et al.\ (2011).  
%\comment{you need to make it plain here that you are describing now this other work, not what we do in this paper; lets rewrite this paragraph together tomorrow.}

In addition to learning about multiple policies, our approach is to learn multiple things about each policy.  Both of these cases can be captured with the notion of  general value functions.  We envision an architecture in which many predictive questions are posed and answered in a generalized form of value function.  Each such function, denoted $\V^i: \SSS \rightarrow \mathbb{R}$, predicts the expected discounted sum of the future readings of some sensor.  The $i$th value function pertains to the sensor readings $r^{(i)}_t$, the policy $\pi^{(i)}$, and the time scale $\gamma^{(i)}$: 

$$\V^i(s) = \mathbb{E}_{\pi^{(i)}} \left[\sum_{k=0}^\infty (\gamma^{(i)})^k r^{(i)}_{t+k+1} \Big|  s_t=s \right].$$

%A natural way to learn many things in parallel on a robot is to consider predictive questions that cover a range of targets at multiple times scales, contingent on a single policy---the behaviour policy. Informally, this can be thought of as learning to predict what will happen next over temporally-extended timescales. Let 

%each question, $\V^{i}, i \in\{1,\ldots, k\}$, be formed using a prediction target, $r^{(i)}$, equal to the current value a sensor or a component of $\phi_t$, for some time scale $\gamma^{(i)}$. 
%The discounted future value of each sensor or feature component is represented as a value function, $\V^{i}(\phi_t)$. The answer to each question can specified by a learned approximate value function, $\hat \V^i$. Each $\hat \V^i$  can then be updated by an independent instance of GTD($\lambda$), given a shared, global feature vector, $\phi_t$.  This simple on-policy setup can be used to construct thousands of distinct questions from a single stream of experience.

%We can dramatically scale this multi-prediction approach for online data by considering off-policy questions.
%Typically, off-policy learning methods are used to learn an optimal control policy given data generated by some alternate behaviour policy, as in Q-learning. 
Off-policy methods can  be used to learn approximate answers $\V_\theta^{(i)}$ 
to predictive questions in the form of such value functions.
Questions about what will happen to robot {\em if} it follows a behaviour different from its current behaviour, for example, `what would be the effect on the rotational velocity, if my future actions consisted of clockwise rotation'. 
%, with a constant probability of terminating on each step. 
Policy-contingent questions substantially broaden the knowledge that can be acquired by the system and dramatically increases the scale of learning---millions of distinct predictive questions can be easily constructed from the space of policies, sensors and time scales.

\begin{figure}
\begin{centering}
\includegraphics[scale=0.5, angle=0]{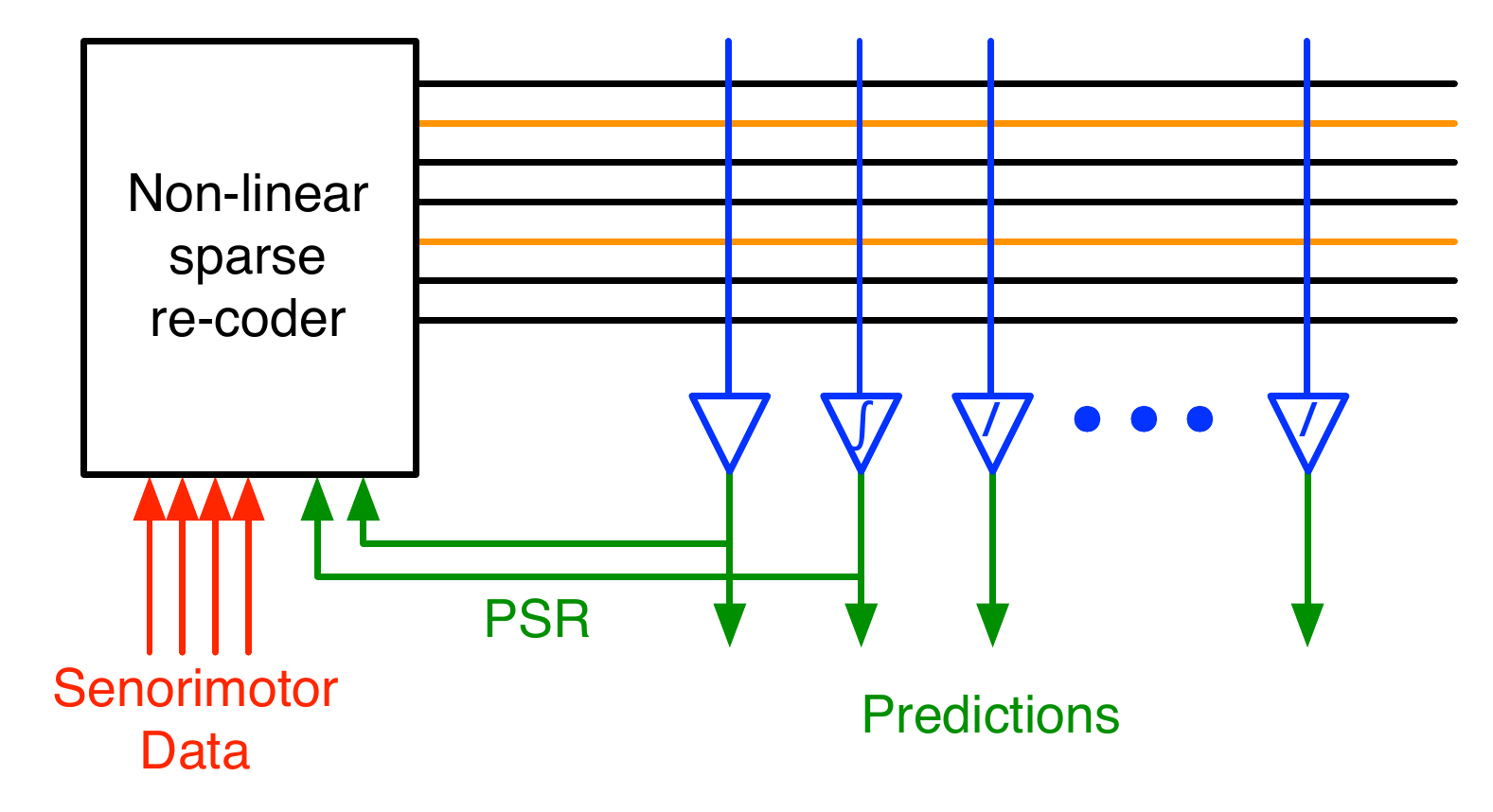}
\caption{\label{horde} The Horde architecture for large-scale off-policy learning. Horde consists of a large set of independent instances of the GTD($\lambda$) algorithm (specified by triangles), updating and making predictions in parallel from a shared set of features. The features are typically a sparse encoding of the raw sensorimotor information and the predictions from the previous time step. The whole system can be operated in parallel and in real time. }
\vspace*{-4mm}
\end{centering}
\end{figure}
Figure \ref{horde} provides a graphical depiction of this parallel learning architecture.
% Each prediction, $\hat \V^{i}$, is learned by an independent instance of GTD($\lambda$) (depicted as triangles), from a shared feature set, produced via a sparse recoding architecture like tile coding. The whole system updates in real-time, learning and making new predictions, from each new sensor packet produced by the robot.    
This architecture, called {\em Horde} by Sutton et al. (2011), has several desirable characteristics. Horde can run in real time, due to the linear computational complexity of GTD($\lambda$). The architecture is potentially scalable because of the distributed nature of off-policy learning, but no experiments were performed to evaluate its ability to learn at scale in practice. The system is modular: the question specification, behaviour policy and the function approximation architecture are completely independent. As depicted in Figure \ref{horde}, the predictions can be used as input to the function approximator. This enables the use of predictive state information (Littman et al., 2002) and learning of compositional predictions, similar to a TD network (Sutton and Tanner, 2005).

\section{Large-scale off-policy prediction on a robot}
\begin{figure}[htbm]
\begin{centering}
\includegraphics[scale=0.2,angle=0]{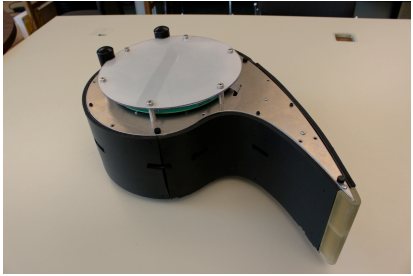} %angle 90 for Adam
\caption{\label{critter} The mobile robot can drive into walls without harm and operate for hours without recharging.}
\vspace*{-0mm}
\end{centering}
\end{figure}

The first question we consider is whether the Horde architecture supports large-scale off-policy prediction in real time on a physical robot. All our evaluations were performed on a custom-built holonomic mobile robot (see Figure \ref{critter}). The robot has a diverse set of 53 sensors for detecting external entities (ambient light, heat, infrared light, magnetic fields, and infrared reflectance) and also its internal status (battery voltages, acceleration, rotational velocity, motor velocities, motor currents, motor temperatures, and motor voltages). The robot can dock autonomously with its charging station and can run continually for twelve hours without recharging.

The raw sensorimotor vector was transformed into features, $\phi_t$, by tile coding.  This produced a binary vector, $\phi_t \in \{0,1\}^n$, with a constant number of 1 features. The tile coder was comprised of many overlapping tilings of single sensors and pairs of sensors.  The tile coding scheme produced a sparse feature vector with $k=6065$ components with 457 features that were ones, including one bias feature whose value was always 1. More details of the feature representation are given in previous work (Modayil, White \& Sutton 2011).

Conducting a fair evaluation presents a challenge for off-policy predictions on a robot; the most direct way to evaluate a prediction about a policy is to follow that policy for a period of time and measure the return.  This direct on-policy test excursion is interspersed into a baseline off-policy behaviour used for learning.  We followed this procedure to evaluate the predictions learned off-policy on the robot, with learning updates being suspended during the test excursions.

To generate behaviour data, the robot was confined to a small two meter square pen, executing one of five actions: $\mathcal{A}=\{$forward, reverse, rotate clockwise, rotate counter clockwise, stop$\}$. A new action was selected for execution every 100ms.  For the baseline learning behaviour, at each time step a random action was selected with a probability of 0.5, otherwise the last executed action was repeated.   Normal execution was interrupted probabilistically to run a test excursion; on average an interruption occurred every 5 seconds. A test excursion consisted of selecting one of five constant action policies and following it  for five seconds. After a test excursion was completed, the robot spent 2 seconds moving to the centre of the pen and then continued its random behaviour policy.   The robot ran for 7.3 hours, visiting all portions of the pen many times. This produced 261,681 samples with half of the time spent on test excursions.
 
We used Horde to learn answers to 795 predictive questions from the experience generated by the behaviour described above. Each question $\V^i$, was formed by combining a $\gamma^{(i)} \in \{0.0,0.5,0.8\}$, a constant action policy $\pi^{(i)}$ from $\{\pi(\cdot, forward) = 1, \pi(\cdot, reverse) = 1, \ldots,  \pi(\cdot, stop)=1\}$, and a prediction target $r^{(i)}$ from one of the 53 sensors. 
Each question was of the form: {\it at the current time $t$, what will be the expected discounted sum of the future values of $r^{(i)}$ if the robot follows $\pi^{(i)}$, with a constant pseudo-termination probability of $1-\gamma^{(i)}$?}. To ease comparison of the predictions across sensors with different output ranges, the values from each sensor were scaled to  the maximum and minimum values in their specifications, so that the observed sensor values were bounded between [0,1].   Each time-step resulted in updates to exactly 159 GTD($\lambda$) learners in parallel (corresponding to the policies that matched the action selected by the behaviour policy). Each question used identical learning parameters:
$\alpha_\theta = 0.1/457$ (457 is the number 
of active features), $\alpha_w = 0.001 \alpha_\theta$, and $\lambda=0.9$. 

The total computation time for a cycle under our conditions was 45ms, well within the 100ms duty cycle of the robot. The entire architecture was run on a 2.4GHz dual-core laptop with 4GB of RAM connected to the robot by a dedicated wireless link. 
  
\begin{figure}[htmb]
\begin{centering}
\includegraphics[scale=0.25]{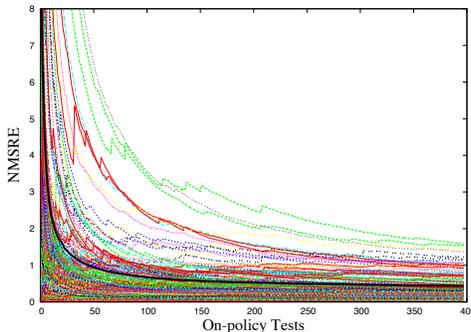} %\angle=270
\vspace*{-10mm}
\caption{\label{fig1} This graph presents the first major result: the first demonstration of learning hundreds policy-contingent predictions at 10 Hz on a consumer laptop.  The x-axis is the number of relevant test excursions observed for each question.  The black heavy stroke line shows the average error over the entire set of questions, and the return error has the typical exponential learning profile. 
 The return errors are normalized by the variance in the return for each question, yielding the percentage of variance unexplained. Several individual curves exhibit non-monotonic shape due to discrepancies between the samples observed under the test excursions and during learning---difficult to avoid while evaluating performance online with individual test excursions.  }
% \TODO{ Adam? Several individual curves have a non-monotonic shape due to discrepancies between the samples observed under the test excursions and during learning---difficult to avoid while evaluating performance online with individual test excursions.} 

\vspace*{-2mm}
%\label{fig1}
\end{centering}
\end{figure}

With the architecture in place to update many off-policy predictions in real time on a robot, we the evaluated on-policy test performance. More precisely, on each test execution, for each of the 159 questions pertaining to the selected test policy, we compared the prediction at the beginning of the test, $\hat{\V}^i(\phi_t)$, with the truncated sample return gathered during the test excursion:
$$\hat g^i_t = \sum_{k=0}^{50} (\gamma^{(i)})^k r^{(i)}_{t+k+1}.$$
Finally, each prediction error was
  normalized by the sample variance of each $\hat g^i_t$ over all starting configurations observed (computed off-line), yielding a normalized mean squared return error (NMSRE)\begin{footnote}{
We use $\overline{x}$ to denote the exponential trace (also called the exponentially weighted average) of samples of $x_t$; this is computed for a fixed time constant $\tau$ by $\overline{x} \equiv \mbox{trace}(x,t) = \frac{1}{\tau} x_t + (1-\frac{1}{\tau}) \mbox{trace}(x,t-1).$}
\end{footnote}
:
\begin{equation}
\textrm{NMSRE}^i_t  = \overline{(\hat{\V}^i - \hat g^i ) ^2}/ \textrm{Var}[\hat g^i].
\end{equation}
The NMSRE represents the percentage of the variance in the returns that remains unexplained by the predictor.  For the questions whose sample returns are constant and thus have a zero sample variance, we define the NMSRE to be one.

Figure \ref{fig1} illustrates our main result: accurate off-policy predictions can be learned, in real time, on a physical robot at scale.  
These predictions were learned from a randomized behaviour policy with a shared feature representation using identical parallel instances of GTD($\lambda$). Note that no question-specific tuning of learning parameters or features was needed.  Another significant result is that no divergence was observed for any question.  Note the  average of the NMSRE for all the questions finished below 1.0: a substantial portion of the variance in the returns is being explained by the predictions. The learning parameters were important---divergence was observed on earlier runs with more aggressive settings (both larger values of $\alpha_\theta$ and smaller values of $\alpha_w$), thus demonstrating that convergence in this off-policy setting is not trivial.

\section{An online measure of off-policy learning\\ progress}
The accuracy of the predictions learned in the previous experiment was evaluated with the return error observed during on-policy test excursions. These tests consume considerable wall-clock time, because for each sample the system must follow the target policy long enough to capture most of the probability mass of the infinite sample return and multiple samples are required to estimate the NMSRE.
% One could restrict testing to a set of special testing policies, subsampled from the larger set of target policies. 
% why is testing performed at all?
%Testing only a subsample of the target policies is problematic because stable learning cannot be guaranteed for all questions.
%?
% Realistically, 
Interspersing on-policy tests to evaluate learning progress places a low limit on both the number of target policies and on the time-scale given by  $\gamma$.

There are other subtle deficiencies with on-policy tests.  The experimenter must choose a testing regime and frequency. Depending on how often tests are executed, there is a trade-off for how often the NMSRE is updated. Changes in the environment and novel robot experiences can cause inaccurate NMSRE estimates if the majority of time-steps are used for training. Testing with greater frequency ensures the estimated NMSRE closely matches current prediction quality, but slows learning.  
%The interaction between the distribution of samples observed during training and the samples encountered during tests can result, noisy estimates of prediction quality (as illustrated in Figure \ref{fig1}).  

In the function approximation setting, we propose instead to estimate the MSPBE. The GTD($\lambda$) algorithm does not minimize the NMSRE, which measures prediction accuracy relative to sample returns, ignoring function approximation error. For an arbitrary question the NMSRE will never go to zero, though it does provide an indication of the quality of the feature representation. The GTD($\lambda$) algorithm instead % as well as every other temporal difference method---
minimizes the MSPBE. Under some common technical assumptions, the MSPBE will converge to a zero error.
% because of the projection operator in the objective function. 
The MSPBE can be estimated in real time during learning, and it provides an up-to-date measure of performance without sacrificing valuable robot data for evaluation.

Using the derivation given by Sutton et al. (2009), we can rewrite this error in terms of expectations:
\begin{flalign} \textrm{MSPBE}&(\theta)=||\V_\theta - \Pi T \V_\theta ||^2_B \\
= ~&||\Pi(\V_\theta - T \V_\theta) ||^2_B \\
= ~&(\Pi(\V_\theta - T \V_\theta))^\top B(\Pi(\V_\theta - T \V_\theta))\\
= ~&(\V_\theta - T \V_\theta)^\top {\Pi}^\top B{\Pi}(\V_\theta - T \V_\theta)\\
= ~&(\V_\theta - T \V_\theta)^\top B^\top\Phi(\Phi^\top B\Phi)^{-1}\Phi^\top B(\V_\theta - T \V_\theta)\\
= ~&(\Phi^\top B(T \V_\theta-\V_\theta))^\top (\Phi^\top B\Phi)^{-1}\Phi^\top B(T \V_\theta-\V_\theta)\\
= ~ & \mathbb{E}_b[\delta\phi]^\top  \mathbb{E}_b[\phi\phi^\top]^{-1} \mathbb{E}_b[\delta\phi]
\end{flalign} 
The GTD($\lambda$) algorithm uses a second set of modifiable weights, $w$, to form a quasi-stationary estimate of the last two terms, namely the product of the inverse feature covariance matrix with the expected TD-update. This leads to the following linear-complexity approximation of the MSPBE:   
\begin{equation}\textrm{MSPBE}(\theta) \approx (\mathbb{E}_b[\delta\phi])^\top w .\label{eq:gtdupdate}\end{equation} 
The expected TD-update term, $\mathbb{E}_b[\delta\phi]$, can be approximated with samples of $\delta_{t} e_t$ (switching from the forward view to backward view), where $e_t$ is the eligibility trace vector. Additionally, the prediction error can be non-zero on samples where the target policy does not agree with the behaviour policy, $\pi^{(i)} (\phi_t,a) \neq b(\phi_t,a)$. The importance sampling ratio, $\frac{\pi^{(i)}(\phi,a)}{b(\phi,a)}$, can be used to account for these effects. This leads to two natural incremental algorithms for sampling the current MSPBE:
\begin{equation} \textrm{MSPBE}_{t,vector} = \overline{\delta e}^\top w_t,\end{equation} and
\begin{equation} \textrm{MSPBE}_{t,scalar} = \overline{\delta e^\top w}.\end{equation}
Here, the exponential traces for both MSPBE$_{t,vector}$ and MSPBE$_{t,scalar}$ are updated on each time step proportionally to $\frac{\pi^{(i)}(\phi,a)}{b(\phi,a)}$.
%The time scale of the trace is set to the time scale of the update to $w$ in the GTD($\lambda$) algorithm ($1.0-\alpha_w$), leading to a parameter free estimate of MSPBE.
The first measure is a more accurate approximation of Equation \ref{eq:gtdupdate}, but the second requires only storing  a single real-valued scalar. 

The first step in evaluating our online estimates of the MSPBE, is to compare them with the exact values of the MSPBE on a simulation domain. We used a simple 7 state Markov chain with an absorbing state on each end, deterministic transitions and episodes beginning in the middle of the chain. Transitioning into the right-side terminal state produced a reward of 1.0, all other transitions incurred 0 reward. We used the inverted feature representation of Sutton et al (2009). 

To determine the validity of our new measures we compared the vector and scalar MSPBE estimates with the true MSPBE (Equation 4) and the expensive sample estimate of the MSPBE (Equation 9). The computation of the true MSPBE requires complete knowledge of the chain MDP. The sample MSPBE requires an incremental estimate of the expected feature covariance matrix and a inverse operation. The sample MSPBE represents the best possible sample-based estimate of the MSPBE; our online measures can not be expected to track the true MSPBE better than the sample MSPBE. In this experiment we used a single instance of GTD with $\alpha_{\theta} = 0.05$, $\alpha_w = 0.1$ and $\lambda = 0.0$ and results were averaged over 100 independent runs. The target policy selected the move-right action with probability 0.95 and a behaviour policy that selected move-right with probability 0.2. 

\begin{figure}
%\begin{centering}
%\includegraphics[width=.51\columnwidth]{EmpiricalReturns}
\includegraphics[width=.5\columnwidth]{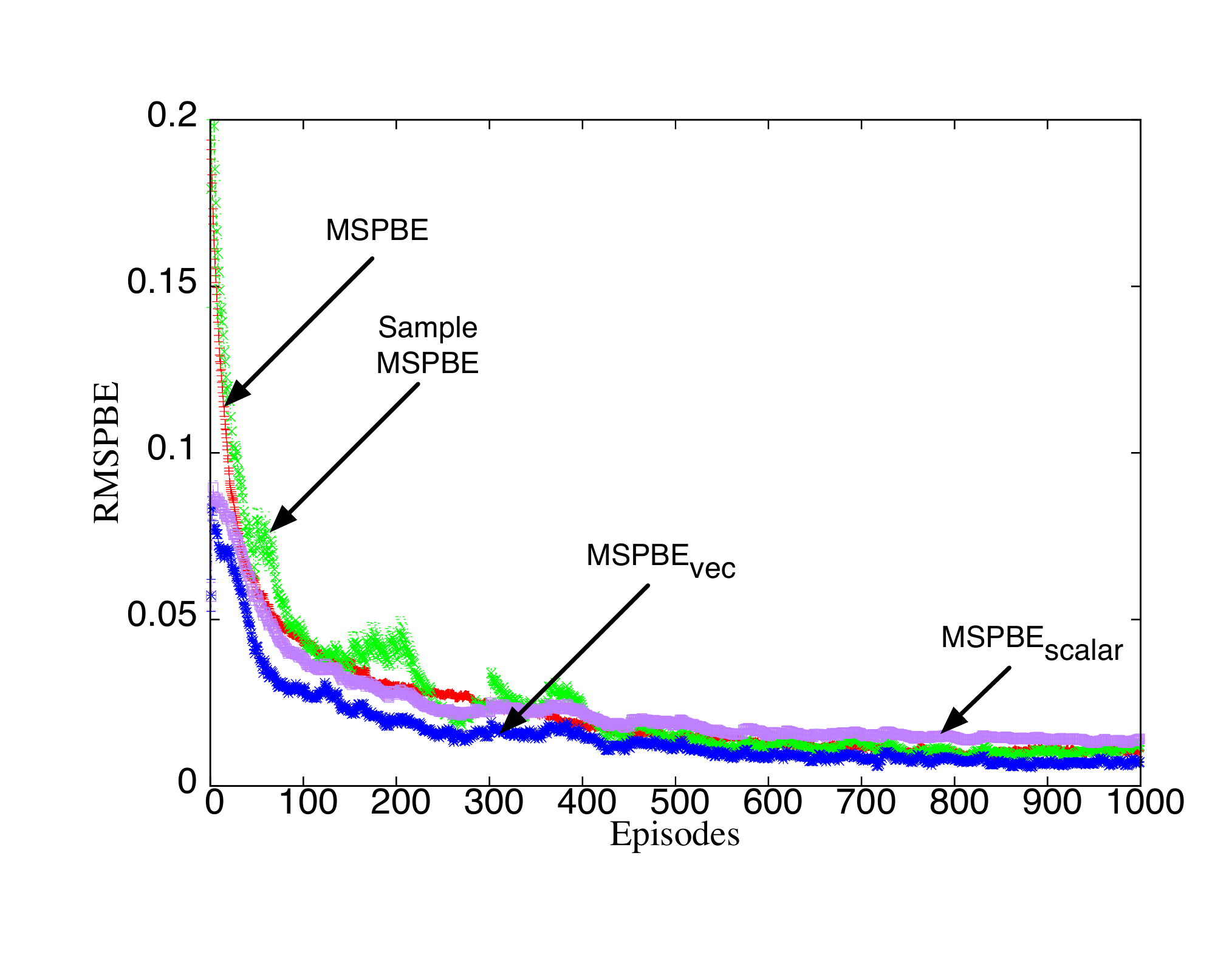}
\includegraphics[width=.5\columnwidth]{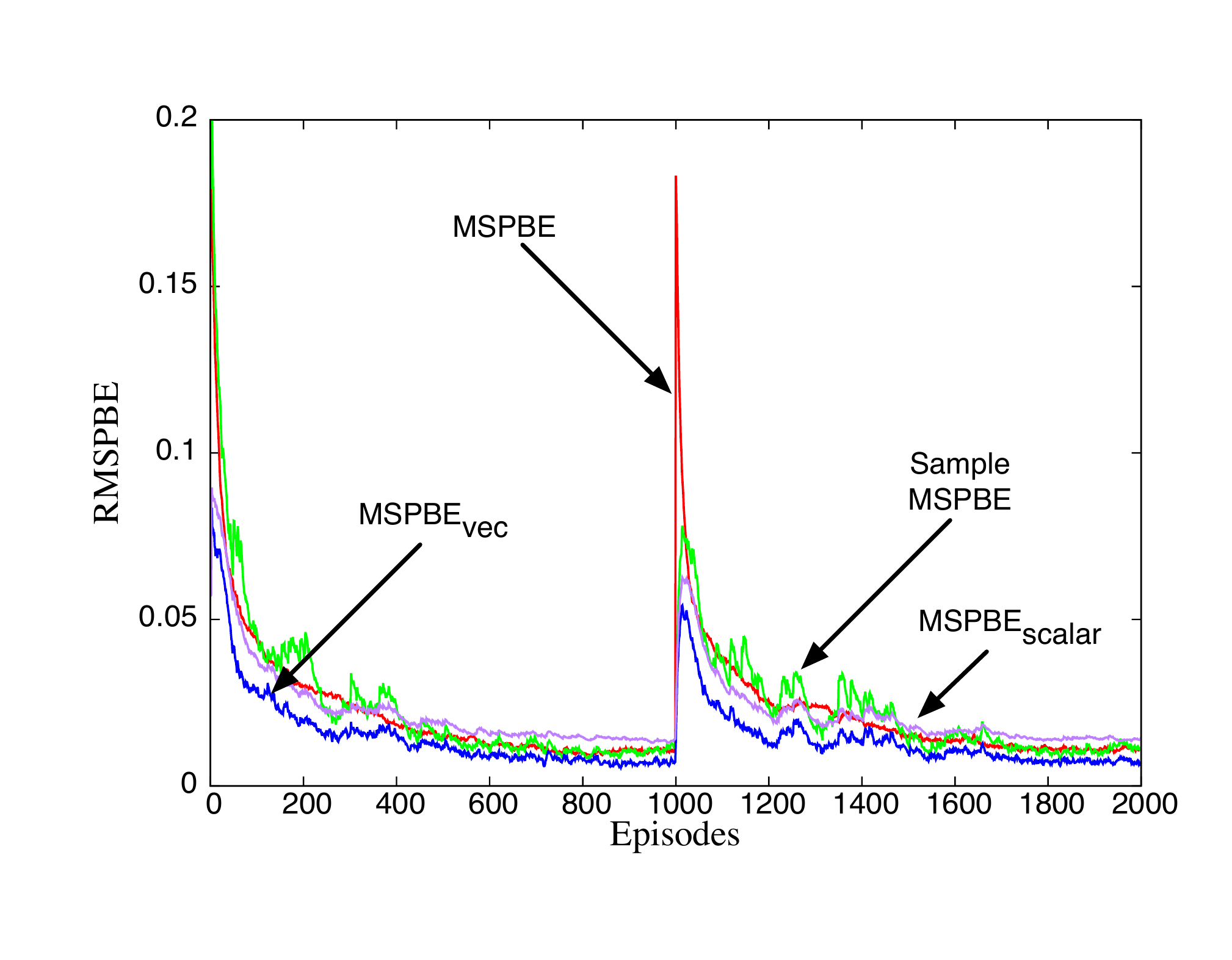}
\vspace*{-12mm}
\caption{\label{chain} Comparison of the online estimates of the MSPBE and the true and sample MSPBE on a simple Markov chain. The left figure figure shows that the measures provide a good estimate of MSPBE matching the profile and magnitude on this simple domain. The right figure compares how the online estimates react to a major change during learning. After 1000 episodes the primary weight vector $\theta$ was set to random values in $[0,1]$. Again, the online estimates track the true and sample MSPBE closely.  }
%\end{centering}
\end{figure}

Figure \ref{chain} (left) illustrates the results of the chain experiment comparison: both our online measures provide an accurate estimate of the MSPBE on a simple chain domain. Figure \ref{chain} (right) compares the effect of a change in the world during learning. After 1000 episodes the $\theta$ weights were set to random values in $[0,1]$. The secondary weights $w$, and traces used in computing the online estimates were not reset. The results depicted in Figure \ref{chain} (right) illustrate that the online estimates, sample MSPBE and the true MSPBE all react similarly to the change.

To evaluate the online MSPBE estimates on the robot, we compare aggregate error curves, averaged over all questions, on tasks where the robot experiences a significant change, similarly to the chain experiment. The robot was run exactly as before, with a subset of the predictions learned ($\gamma = 0.8$), for six hours. This time, the learned weight vector of each question $\theta^{(i)}$, was set to zero after 40000 time steps. This change effectively reinitializes each question and effects the accuracy of all the predictions. In this experiment, we recorded the NMSRE, MSPBE$_{t,vector}$ and MSPBE$_{t,scalar}$ on every time step for 265 questions, except during test excursions. Note that the NMSRE is only updated after a test completes, while the MSPBE measures are updated on every non-test time-step. 

\begin{figure}[htmb]
\begin{centering}

\vspace*{-4mm}
\includegraphics[scale=0.4, angle=0]{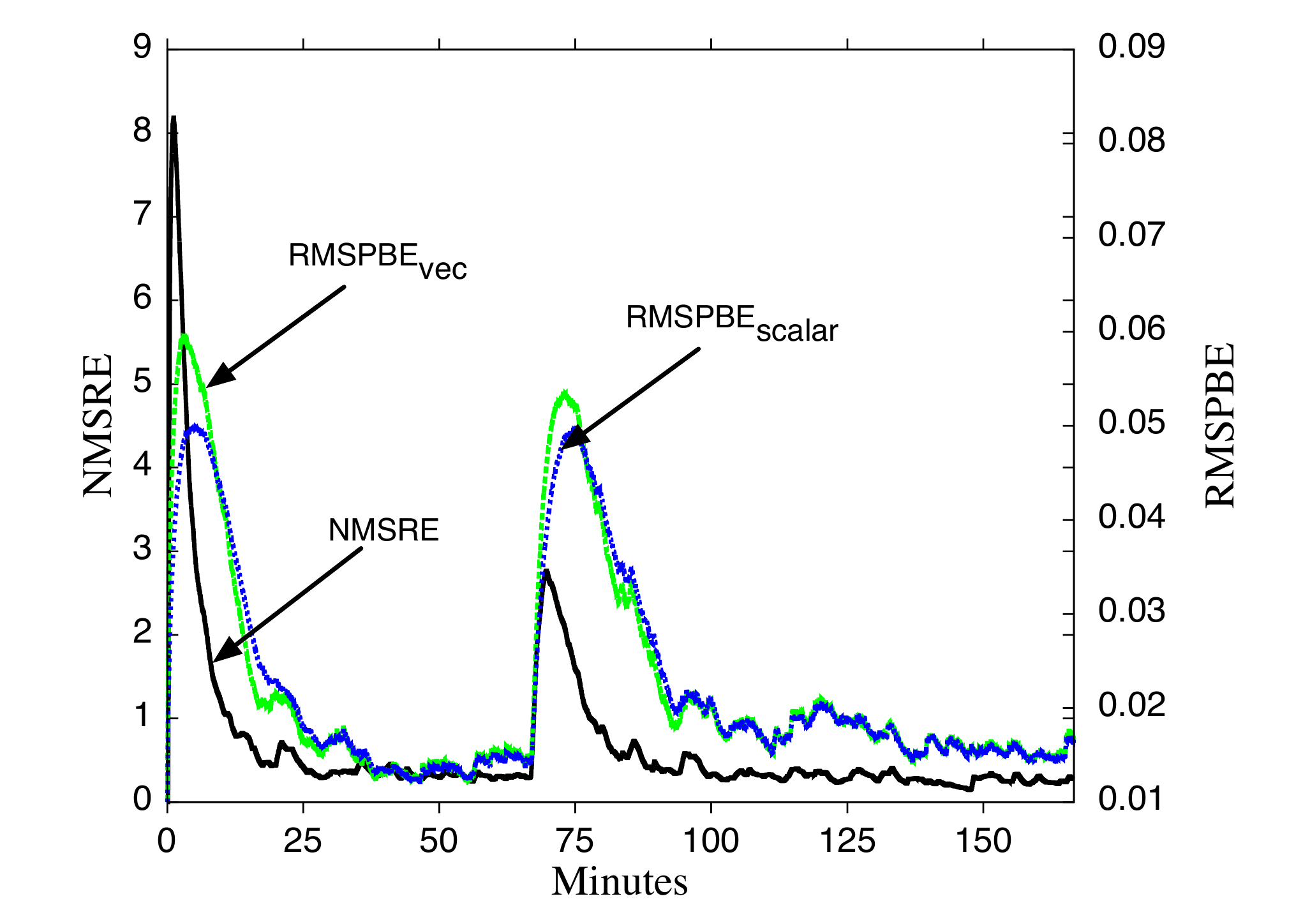}%ICMLFigure2_finalp}%ICMLFigure4} %angle=270
\vspace*{-4mm}
\caption{\label{figure4} This figure compares the NMSRE, with two estimates of the MSPBE, averaging performance over 265 predictive questions. This figure illustrates several important points that validate the MSPBE as a useful error measure. The MSPBE measures as computed off-policy have the same shape as the NMSRE which requires an on-policy excursion in the behaviour. The MSPBE estimates converge more slowly than the NMSRE indicating that although test performance is not improving, learning is still occurring. The MSPBE estimates react more slowly to the change because the secondary weights $w$ where not also reset; their incorrect values affect the estimates of the MSPBE long after the change. The MSPBE measures react quickly to a change in the environment. This experiment shows that the scalar estimate of the MSPBE performs almost identically to the vector version. This graph illustrates a significant result: off-policy learning progress can be efficiently estimated online, without time consuming on-policy tests. Additionally that the NMSRE requires twice as twice as much wall-clock time as the MSPBE estimates. }
\label{errorCompare}
\end{centering}

\end{figure}

Figure \ref{errorCompare} compares the convergence profile and reaction to change of the three error measures in terms of training time. As in the chain experiments, all three measures react quickly to the change. Note that both MSPBE estimates are initially at zero, as the vector $w$ takes time to adapt to a useful value.  Finally, note that the MSPBE$_{t,vector}$ and MSPBE$_{t,scalar}$ exhibit very similar trends, indicating that the Bellman error can be estimated with minimal storage requirements.
       
%\begin{figure}
%\includegraphics[scale=0.3, angle=270]{figs/ICMLFigure3}
%\caption{\label{figure3} Descript.}
%\vspace*{-4mm}
%\end{figure}

\section{Large-scale off-policy prediction, with many target policies}
Free from the limitations of physically performing test excursions to evaluate predictions, we can learn about a much larger set of questions. In this section, we consider scaling the number of target policies and prediction time scales (magnitude of $\gamma$).

To increase the space of target policies, and still maintain a small set of finite actions, we consider discrete-action linearly parametrized Gibbs policy distributions:
$$\pi_u(a) = \frac{ \exp({- u^\top \Psi_a  }) } { \sum_{a'\in \mathcal{A}} \exp({- u^\top \Psi_{a'} })}$$
where $u$ is a vector of policy parameters.
% and $\tau$ is a temperature parameter that controls exploration.
 The feature vector for each action, $\Psi_a \in \mathbb{R}^{n|\mathcal{A}|}$, has a copy of $\phi_t$ as a subvector in an otherwise zero vector; and for each action the copy is offset by $n$ so that $a\neq a' \implies \Psi_a^\top \Psi_{a'}=0$.  Random policies are generated by selecting 60 components of $u$ at random and assigning each a value independently drawn from the uniform distribution over $[0,1]$. 

In this final experiment, we tested how well our architecture scales in the number of target policies. The robot's behaviour was the same as before, but now learning was enabled on every step of the 7 hours experience. The questions were formed by sampling $\gamma$ values from $\{0.0, 0.5, 0.8,0.95\}$, reward from the full set of sensors with 1000 randomly generated policies. The value of $\gamma~=~0.95$ corresponds to a 2 second prediction and would require over 30 seconds to accurately evaluate using the NMSRE. 
The 1000 questions, evaluated according to MSPBE$_{t,scalar}$, were learned with a cycle time of $85$ms on a 4-core desktop computer with 16 G of Ram; satisfying our real-time requirement of $100$ms.  

\begin{figure}
\begin{centering}
\vspace{-3mm}
\includegraphics[scale=0.6]{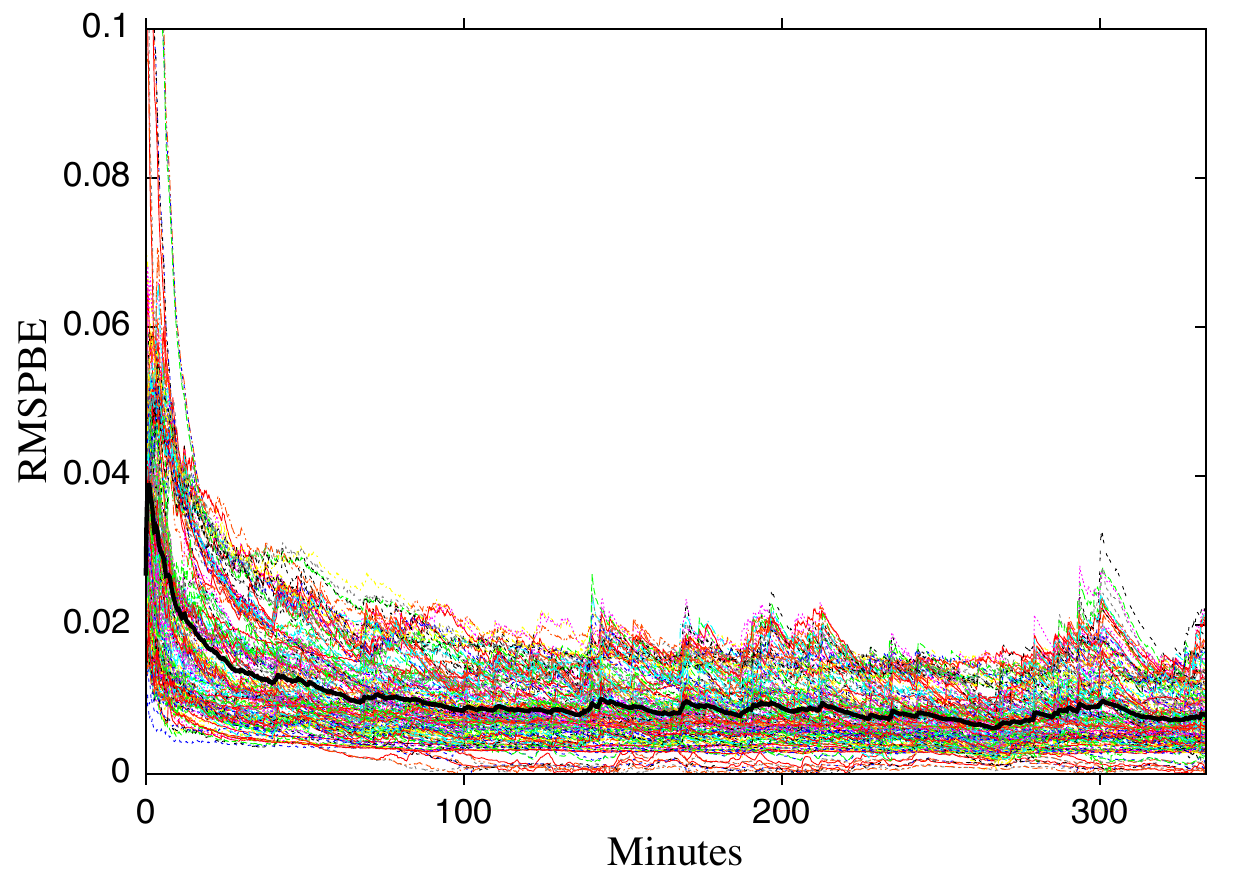}%ICMLFigure3_final3} %angle=270
%\vspace{-6mm}
\caption{\label{figure2} Scaling off-policy learning to 1000 robot policies.
The estimated mean squared projected Bellman error for off-policy predictions for $1000$ distinct randomly generated policies. The heavy stroke black line denotes the average estimated error over the full set of questions. The curves provide a clear indication of learning progress for each prediction.
This result provides a clear demonstration of the significance of estimating the MSPBE incrementally during learning. Massively scaling the number of policies on which to condition predictions is only possible with an online performance measure.
}
\label{scaleLearning}
\vspace*{-4mm}
\end{centering}
\end{figure}

Figure \ref{scaleLearning} presents the results of this experiment, namely that learning the temporally-extended consequences of many (untestibly many) different behaviours is possible in real time.  The learning progress is measured by the MSPBE, which by the results in the previous section will be strongly coupled to on-policy prediction errors.  Note that the ability to monitor learning progress across so many different behaviours is only possible due to the availability of the MSPBE. By acquiring many predictions about many different courses of behaviour, the robot can acquire detailed partial models of the dynamics of its environmental interaction.  

% \section{Discussion}
% Given space we can discuss and present several extensions and additional results.
% \begin{itemize}
% \item additional expressiveness achieved by considering state dependent $\gamma$ and outcome rewards ($z$)
% \item constructing time series of future rewards
% \item Feeding the predictions back into the state
% \item off-policy control results (results borrowed from horde paper)
% \end{itemize}

%%\begin{figure}
%%\includegraphics[scale=0.3, angle=270]{figs/ICMLFigure5}
%%\caption{\label{figure4} Descript.}
%%\vspace*{-4mm}
%%\end{figure}

\section{Related Work}
%\begin{itemize}
%\item D. Pierce and B. Kuipers. 1997
%\item options
%\item Doina - recognizers/ importance sampling for off-policy learning
%\item

%Intrinsically Motivated Reinforcement Learning (playroom)

%\item Thrun's life-long learning 
%\item 

%TD-nets and PSRs

%\item T-PSRs
%\item robotics (Atkeson and Schaal, Abbeel and Ng, Peters)
%\item bayesian approaches
%\end{itemize}

Many of the ideas in this paper have precursors in the literature.  The idea of policy-contingent predictions was developed along with the options framework for temporal abstraction (Sutton, Precup \& Singh, 1999).  Learning off-policy under function approximation was developed by importance sampling (Precup et al., 2006) in an approach that runs online, but can exhibit exponentially slow learning progress.  Learning about many different policies can also support active exploration, an idea that has been explored in related work on curiosity-based learning (Singh et al., 2005).

%The algorithm for faster learning with gradient-based temporal-difference algorithms is given in a set of recent papers.  
%Related work even suggested that these ideas scaled to off-policy learning, but the scale was never demonstrated.

The idea of building models from data has been explored, but not in the off-policy real-time setting. Thrun and Mitchell (1995) showed learning of sensor models offline.  Many Bayesian approaches for online state estimation in robotics (e.g., Thrun et al., 2005) process a vast volume of observations in real time, but they do not learn system dynamics online.  A Kalman filter (Kalman, 1965)  can be viewed as  learning an adaptive model of dynamics online, but it is only appropriate for small models with well-understood dynamics.  Atkeson and Schaal (1997) showed learning of small one-time-step models.  Kober and Peters (2011) showed on-policy episodic learning on a robot.

%The system architecture come from (Sutton et al., 2011) along with ideas from predictive state representations (Littman et al., 2002) and temporal-difference networks (Sutton \& Tanner, 2005), but these works are small demonstrations that do not demonstrate learning at this scale. 

A recent online spectral approach  (Boots, Siddiqi \& Gordon, 2011) finds small predictive state models for robotics.  They use a sophisticated incremental method for making multiple action-conditional predictions, but it is unclear if this approach can operate under real-time constraints.  The previous work that introduced the Horde architecture (Sutton et al., 2011) also demonstrated parallel off-policy learning on a robot.  Their work was suggestive but did not demonstrate that the approach scaled in practice.  Each experiment required a unique parameter set, function approximation scheme, and behaviour policy; this overhead is not practical for learning thousands of predictions.   This paper shows that a large set of diverse, policy-contingent predictions can be learned using a shared feature set, common learning parameters, and data generated by a single random behaviour policy.

% Another scaling issue involves evaluation---determining the accuracy of predictions about policies never followed by the robot.

\section{Conclusions and future work}
We provided the first demonstrations of large scale off-policy learning on a robot. We have shown that gradient TD methods can be used to learn hundreds of temporally-extended policy-contingent predictions from off-policy sampling. To achieve this goal required resolution of several challenges unique to the off-policy setting. Most significantly we have developed on online estimate of off-policy learning progress based on the Bellman error that does not increase the computational complexity of the horde architecture, can be sampled without interrupting learning and has good correspondence with the traditional mean squared prediction error. The addition of policy contingent, what-if, questions dramatically increases the scope and scale of questions that can be learned by horde providing further evidence of the significance of horde for life-long learning.

Our experiments, while on a robot, are limited but there are several immediate directions for future work. The questions learned here are predictive questions about a policy, general value functions can also support learning control policies using greedy-gq (Maei 2011) a control variant of GTD($\lambda$) with the same linear complexity. Predictive accuracy improvements should be achieved by employing adaptive behaviour policies (e.g., curiosity) and more powerful function approximators, while significantly increased scaling (more predictions and larger feature vectors) can be obtained with more computational resources. 
% 
%%Estimating performance online is critical for scaling off-policy learning beyond a handful of target policies, as demonstrated in this paper. 
% We have shown that off-policy learning from sensorimotor streams has considerable parallel scalability. 
% Finally, the robustness and scale of this method suggest it can support studies into adaptive behaviour policies, representation search, and off-policy control learning.

% In the unusual situation where you want a paper to appear in the
% references without citing it in the main text, use \nocite
%\nocite{langley00}

%\bibliography{rtbd}
%\bibliographystyle{icml2012}
\parindent=0pt
\def\hangin{\hangindent=0.2in}

\section{References}

%\footnotesize

%\small
\parskip=2pt

\hangin
Atkeson, C.~G.,  Schaal, S. (1997). Robot learning from
demonstration. In {\it Proc.\ 14th Int.\ Conf.\ on Machine Learning}, pp.~12--20.

\hangin
Baird, L.~C. (1995).
\newblock Residual algorithms: {R}einforcement learning with function
  approximation.
\newblock In {\it Proc.\ 
  12th Int.\ Conf.\ on Machine Learning}, pp.~30--37.
	 
\hangin
Boots, B., Siddiqi, S., Gordon, G. (2011).
An online spectral learning algorithm for
partially observable nonlinear dynamical systems. 
In {\it Proc.\ Conf.\ of the Association for the Advancement of Artificial Intelligence}.

%\hangin
%Bradley, J.~K., Kyrola, A., Bickson, D.,  Guestrin, C. (2011).
%Parallel coordinate descent for L1-regularized loss minimization. 
%In {\it Proc.\ Int.\ Conf.\ on Machine Learning}.

%\hangin
%Chen, W., Song, Y., Bai, H., Lin, C.,  Chang, E.~Y. (2011).
%Parallel spectral clustering in distributed systems.
%{\it IEEE Trans.\ Pattern Analysis and Machine Intelligence 33}(3):568--586.

%\hangin
%Gonzalez, J., Low, Y., Guestrin, C., O'Hallaron, D. (2009).
%Distributed parallel inference on large factor graphs.  
%In {\it Proc.\ Uncertainty in Artificial Intelligence}.

\hangin
Hsu, D., Karampatziakis, N., Langford, J., Smola, A.~J. (2011).
Parallel Online Learning. 
In {\it The Computing Research Repository}.

\hangin
Kalman, R.~E. (1960).
A new approach to linear filtering and prediction problems. 
{\it Trans.\ ASME, Journal of Basic Engineering 82}:35--45.

\hangin
Kober, J., Peters, J. (2011).
Policy search for motor primitives in robotics. 
{\it Machine Learning 84}:171--203.

\hangin
Kolter, J.~Z. (2011).
The fixed points of off-policy TD.  
In {\it Advances in Neural Information Processing Systems 24}.

\hangin
Littman, M.~L., Sutton, R.~S., Singh, S. (2002). 
Predictive representations of state. 
In {\it Advances in Neural Information Processing Systems 14}, pp.~1555--1561.
 
\hangin
Maei, H.~R. (2011).
Gradient Temporal-Difference Learning Algorithms. PhD thesis, University of Alberta.

\hangin
Modayil, J., White, A., Sutton, R.~S. (2012).
Multi-timescale Nexting
in a Reinforcement Learning Robot.
{\it Proc.\ 12th  Int.\ Conf.\ on Adaptive Behaviour}.

\hangin
Precup, D., Sutton, R.~S., Paduraru, C., Koop, A., Singh, S. (2006).
Off-policy learning with recognizers. 
In {\it Advances in Neural Information Processing Systems 18}.

\hangin
Quadrianto, N., Smola, A., Caetano, T., Vishwanathan, S.~V.~N.,  Petterson, J. (2010).
Multitask learning without label correspondences. 
In {\it Advances in Neural Information Processing Systems 23}, pp.~1957--1965.

\hangin
Singh S., Barto, A.~G., Chentanez, N. (2005).
Intrinsically motivated reinforcement learning.
In {\it Advances in Neural Information Processing Systems 17}, pp.~1281--1288.

\hangin
Sutton, R.~S. (1988).
Learning to predict by the method of temporal differences.
{\it Machine Learning 3}:9--44.

\hangin
Sutton, R.~S., Barto, A.~G. (1998).
{\it Reinforcement Learning: An Introduction}.
MIT Press.

\hangin
Sutton, R.~S., Precup D., Singh, S. (1999).
 Between {MDP}s and semi-{MDP}s: A framework for temporal
abstraction in reinforcement learning. 
 {\it Artificial Intelligence 112}:181--211.

\hangin
Sutton, R.~S., Tanner, B. (2005).
Temporal-difference networks.  
In {\it Advances in Neural Information Processing Systems 17}, pp.~1377--1384.

\hangin
Sutton, R.~S., Maei, H.~R., Precup, D., Bhatnagar, S., Silver, D., Szepesv\'ari, Cs.,  Wiewiora, E. (2009).
Fast gradient-descent methods for temporal-difference learning with linear function approximation. 
In {\it Proc.\ 26th Int.\ Conf.\ on Machine Learning}.

\hangin
Sutton, R.~S., Modayil, J., Delp, M., Degris, T., Pilarski, P.~M., White, A.,
   Precup, D. (2011).
Horde: A scalable real-time architecture for learning knowledge from
  unsupervised sensorimotor interaction.
{\it Proc.\ 10th Int.\ Conf.\ on Autonomous Agents and Multiagent Systems}.

\hangin
Talvitie, E.,  Singh, S. (2011). 
Learning to make predictions in partially observable environments without a generative model.
{\it Journal of Artificial Intelligence Research 42}:353--392.

\hangin
Thrun, S., Burgard, W.,  Fox, D. (2005).
{\it Probabilistic Robotics}. MIT Press.

\hangin
Thrun, S., Mitchell, T. (1995).
Lifelong robot learning. {\it Robotics and Autonomous Systems}.

\end{document}